\documentclass[journal]{IEEEtai}

\usepackage[colorlinks,urlcolor=blue,linkcolor=blue,citecolor=blue]{hyperref}

\usepackage{color,array}

\usepackage{graphicx}

\usepackage{amsmath, amssymb}
\usepackage{cleveref}
\usepackage{threeparttable}
\usepackage[nolist]{acronym}

\usepackage{tabularray}
\UseTblrLibrary{booktabs}
\SetTblrInner{rowsep=1pt,colsep=3pt}

\newcommand{\ie}{i.\,e.,}

\newcommand{\eg}{e.\,g.,}

\newcommand{\acr}[1]{\acs{#1} (\acl{#1})\acused{#1}}

\usepackage[backend=biber,style=ieee,natbib=true,maxcitenames=2,mincitenames=1,doi=true,isbn=false,url=true,eprint=false]{biblatex}
\begin{document}

% \title{\acs{TFMPathy}: Tabular Foundation Model for Privacy-Aware, Generalisable Empathy Detection from Videos}
% \title{Privacy-Preserving Empathy Detection in Human-Robot Interaction Videos}
\title{Privacy-Preserving Empathy Detection in Video Interactions}
% \title{Privacy-Aware and Generalisable Empathy Detection from Videos Using Tabular Foundation Model}
% \title{Tabular foundation model to detect empathy from visual cues}

\author{Md~Rakibul~Hasan,~\IEEEmembership{Graduate~Student~Member,~IEEE,}
Md~Zakir~Hossain,
Aneesh~Krishna,~\IEEEmembership{Member,~IEEE,}
Shafin~Rahman,
and~Tom~Gedeon,~\IEEEmembership{Senior~Member,~IEEE}
\thanks{M R Hasan, A Krishna and T Gedeon are with School of Electrical Engineering, Computing and Mathematical Sciences, Curtin University, Bentley, WA 6102, Australia.}% <-this % stops a space
\thanks{M Z Hossain is with The Australian National University, Canberra, ACT 2600, Australia.}
\thanks{S Rahman is with North South University, Dhaka 1229, Bangladesh.}
% \thanks{M R Hasan is also with BRAC University, Dhaka 1212, Bangladesh.}
\thanks{T Gedeon is also with The Australian National University, Canberra, ACT 2600, Australia, and University of ÓBuda, 1034 Budapest, Hungary.}
\thanks{E-mail: \{rakibul.hasan, a.krishna, tom.gedeon\}@curtin.edu.au, zakir.hossain@anu.edu.au, shafin.rahman@northsouth.edu}
\thanks{Corresponding author: M R Hasan}
}
% \thanks{This paragraph of the first footnote will contain the date on which you submitted your paper for review. It will also contain support information, including sponsor and financial support acknowledgment. For example, ``This work was supported in part by the U.S. Department of Commerce under Grant BS123456.'' }
% \thanks{The next few paragraphs should contain the authors' current affiliations, including current address and e-mail. For example, F. A. Author is with the National Institute of Standards and Technology, Boulder, CO 80305 USA (e-mail: author@boulder.nist.gov).}
% \thanks{S. B. Author, Jr., was with Rice University, Houston, TX 77005 USA. He is now with the Department of Physics, Colorado State University, Fort Collins, CO 80523 USA (e-mail: author@lamar.colostate.edu).}
% \thanks{T. C. Author is with the Electrical Engineering Department, University of Colorado, Boulder, CO 80309 USA, on leave from the National Research Institute for Metals, Tsukuba, Japan (e-mail: author@nrim.go.jp).}
% \thanks{This paragraph will include the Associate Editor who handled your paper.}}

% \markboth{Journal of IEEE Transactions on Artificial Intelligence, Vol. 00, No. 0, Month 2026}
% \markboth{IEEE Transactions on Artificial Intelligence, Vol. 00, No. 0, Month 2026}
\markboth{UNDER REVIEW}
{M R Hasan \MakeLowercase{\textit{et al.}}: Privacy-Preserving Empathy Detection in Video Interactions}

\IEEEpubid{This work has been submitted to the IEEE for possible publication. Copyright may be transferred without notice.}

\maketitle

\begin{abstract}
Detecting empathy from video interactions has emerging applications, yet raw videos that could be used for training AI models are rarely available due to privacy and ethical constraints. Public benchmarks are consequently released only as pre-extracted features, creating a privacy-constrained learning regime whose privacy-utility trade-off is poorly characterised. We formalise three levels of privacy for video-based behavioural prediction -- \emph{no privacy} (raw video), \emph{partial privacy} (temporal visual features such as facial landmarks, action units and eye gaze) and \emph{strong privacy} (summary statistics of those features)-- and ask whether strong, subject-generalisable empathy detection is achievable at the strong-privacy level. We propose \acs{TFMPathy}, instantiated with two recent \acp{TFM} (TabPFN v2 and TabICL), under both in-context learning and fine-tuning paradigms. On a public human-robot interaction benchmark, \acs{TFMPathy} achieves strong utility under strong privacy, outperforming established baselines by a substantial margin. To assess robustness and facilitate fair, safe deployment, we introduce a cross-subject evaluation protocol that was previously lacking in this benchmark. Under this protocol, \ac{TFM} fine-tuning improves generalisation capacity substantially (accuracy: $0.590 \rightarrow 0.730$; \acs*{AUC}: $0.564 \rightarrow 0.669$). Aggregating temporal features into summary statistics also suppresses subject-specific and demographic cues, aligning \acs{TFMPathy} with data-minimisation principles. \acs{TFMPathy}, therefore, offers a practical route to building AI systems that depend on human-centred video when governance, consent or institutional policies restrict the sharing of raw video. Code will be released upon acceptance at \url{https://github.com/hasan-rakibul/TFMPathy}.
\end{abstract}
% with a clear privacy-utility trade-off: Prior work on this setting has established classical tree-based models as the state of the art. Motivated by recent advances of large-scale foundation models, we investigate whether \acp{TFM} can recover predictive utility in this setting.

\begin{figure}[t!]
    \centering
    \includegraphics[trim=0 0 37 0, clip, width=1\linewidth]{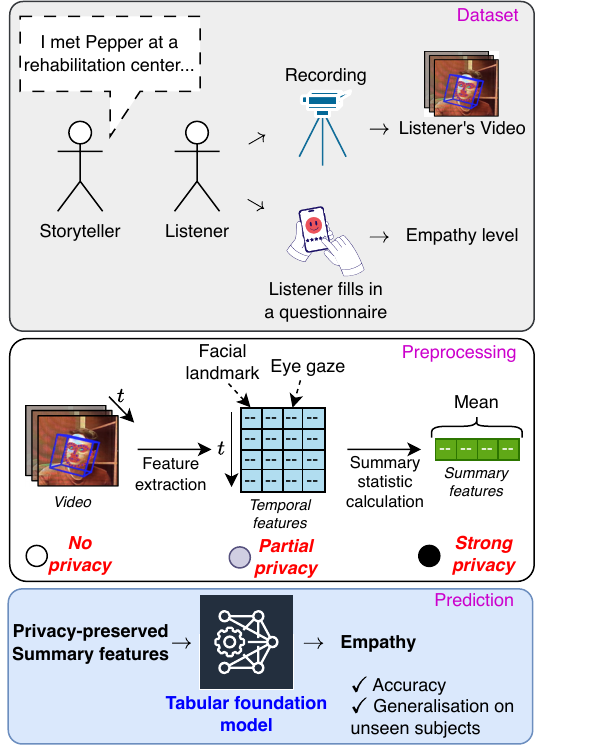}
    \caption{Overview of the task and motivation. We study privacy and fairness in detecting empathy of subjects listening to narrative storytelling, where the listeners self-report their empathy levels. Three levels of privacy are considered. Using raw video provides \textit{no privacy}. Using frame-level visual features, such as eye gaze and facial landmarks, provides \textit{partial privacy}, as these features retain enough information to enable face reconstruction and re-identification. Using summary statistics of those features provides \textit{strong privacy} under our practical, data-minimisation framing, and is the regime in which we develop \acfp{TFM}-based pipeline for empathy prediction. Results show that our strong privacy approach with \acs{TFMPathy} provides strong performance under an established evaluation protocol and retains competitive utility in our new cross-subject generalisation protocol.}
    \label{fig:premise}
\end{figure}

\begin{IEEEImpStatement}
As AI systems are increasingly developed for human-centred video tasks, ensuring that such systems respect privacy and treat individuals fairly is essential for public trust and safe deployment. Further, due to privacy constraints and data-protection regulations such as GDPR, raw videos are not released in many public benchmarks, necessitating the development of AI methods without raw videos. We address this necessity through the emerging task of empathy detection in video interactions, showing that empathy can be reliably predicted from summary statistics of behaviours in video rather than from actual video. This work, therefore, enables us to study and improve empathic interaction without exposing human subjects to re-identification, without breaching data protection laws, and without the energy cost of training large video models. The work also tests whether predictions hold for people unseen during training -- a fairness check for deployment in diverse populations, and a step toward trustworthy AI that promotes privacy and fairness.
\end{IEEEImpStatement}

\IEEEpubidadjcol


\acresetall

\section{Introduction}

% empathy, its detection, and groundwork for empathy towards robots
\IEEEPARstart{E}{mpathy}, as defined in social psychology \citep{davis1996empathy,hoffman2000empathy}, refers to an individual's capacity to perceive, comprehend and vicariously experience the emotional state of others. This interpersonal skill has practical significance in a variety of interaction contexts such as \acp{HRI}, clinical communication and educational settings. A training intervention to nurture this skill requires \textit{assessment}\footnote{We use the terms `assessment', `measurement', `detection' and `prediction' interchangeably.} of empathy. Current assessment approaches rely on self-reported questionnaires or labour-intensive manual analysis by external observers \citep{bas2020empathy}. Automated empathy detection through \ac{ML} techniques has gained momentum over the past decade \citep{hasan2025empathy}. While the majority of this progress has occurred within natural language processing contexts, \ie{} detecting empathy in textual content, empathy detection from video data remains relatively under-explored \citep{hasan2025empathy}. Towards this direction, this paper measures empathy from videos of human interactions.
Specifically, we measure the level of empathy elicited by storytelling, \ie{} the extent to which a listener demonstrates empathy towards others portrayed in a story \citep{mathur-acii-hri,keen2006theory}.

In this domain of empathy elicited in storytelling videos, \citet{mathur-acii-hri} released a public benchmark featuring empathy elicited by a robot storyteller. Named as the \ac{HRI} dataset in this paper, this dataset involves human subjects attending to a robot telling different stories that may elicit empathy. Although there are multiple sessions per subject, prior work has not evaluated scenarios in which the same subject's data may appear in both the training and test sets. 
This is both a statistical concern -- empathy is subjective, so subject overlap inflates apparent generalisation -- and a fairness concern, because per-subject memorisation disadvantages users whose patterns are under-represented in training.
To address this, we conduct a subject-independent evaluation scheme on this benchmark, ensuring no subject is shared between the training and test sets. 

Video-based empathy detection sits within a broader social context in which raw video of human subjects is treated as sensitive personal data. Under modern data protection regulations such as the General Data Protection Regulation (GDPR), the \emph{data minimisation} principle states that personal data should be ``adequate, relevant and limited to what is necessary'' \citep{it2020eu}. The National Institute of Standards and Technology (NIST) Privacy Framework outlines five core functions (Identify, Govern, Control, Communicate, and Protect) and emphasises that managing privacy risks arising from data processing is critical for safeguarding individuals \citep{nist2020nist}. Ensuring privacy and fairness is also foundational towards the broader goal of \emph{trustworthy AI} \citep{pfau2026engineering}. For these reasons, the release of raw human-centred video is discouraged beyond what is strictly required. Reflecting this, many public affective computing benchmarks, such as \ac{HRI} empathy benchmark \citep{mathur-acii-hri, spitale-jhri-hri}, release only pre-extracted behavioural features rather than raw video footage. Since such feature-only representation is an emerging norm in video-based affective computing research, we need to develop methods for real-world deployment that use only such feature data.

Critically, however, not all feature-only representations are equally privacy-preserving. Frame-level facial landmarks, action units and gaze vectors retain sufficient geometric and dynamic information to support face reconstruction \citep{di2018gpgan} and to leak attributes such as identity, gender, age and emotional state \citep{kroger2020what, mase2023facial}. We therefore distinguish between two feature-only regimes: \emph{partial privacy}, in which temporal sequences of visual features are exposed, and \emph{strong privacy}, in which only temporal summary statistics (\eg{} per-feature mean, median, standard deviation and autocorrelation) are exposed.
Summary statistics are a lossy, non-invertible projection of the underlying time series and therefore discard the fine-grained temporal signatures that per-subject biometrics rely on. \Cref{fig:overview} illustrates the three levels -- \emph{no privacy} (raw video), \emph{partial privacy} (temporal features), and \emph{strong privacy}\footnote{We use the term strong privacy in this paper to denote a summary-level representation that substantially reduces access to raw appearance and fine-grained temporal trajectories. This should not be interpreted as anonymity, differential privacy, cryptographic privacy, or a guarantee against all identity, demographic, behavioural, or health-related inference.} (summary statistics) -- that structure the rest of this paper.

% I think this para can be moved to discussion section if the intro appears too long
This graded notion of privacy has a second benefit: it constrains the extent to which a predictive model can exploit identifying cues as shortcuts. Face-based affect recognition systems are known to exhibit demographic disparities across race, gender and age \citep{buolamwini2018gender, xu2020investigating, kim2021age, booth2021integrating}. Adding the face/video modality has been shown to \emph{increase} bias in multimodal emotion pipelines relative to text-only baselines \cite{schmitz2022bias}. Our method, therefore, avoids using raw face/video and instead uses summary statistics, which are less biased due to not having access to the videos. Our method may also reduce some visual demographic leakage, noting that demographic fairness cannot be confirmed without demographic subgroup annotations.

% Empathy detection varies widely due to seemingly similar affective computing tasks like emotion recognition in many ways. First, empathy is subjective, so the annotated dataset
% Empathy is often reciprocal (\ie{} both parties being empathic to each other) \citep{calvo2014reciprocal}, so providing empathy increases the chance of receiving empathy.

% already covered in the earlier paragraph
% empathy from video features --> tabular data
% Being an emerging area of research, the number of publicly available video datasets is limited \citep{hasan2025empathy}. 
% Datasets of raw video footage featuring people's activity are difficult to make public due to privacy and ethical concerns. Consequently, the majority of these existing public datasets, such as human-robot interactions \citep{mathur-acii-hri,spitale-jhri-hri}, are released as extracted feature vectors (\ie{} tabular data) rather than raw video footage. Since privacy and ethical concerns are likely to persist, one key challenge in this domain is to develop \emph{privacy-preserving} empathy detection system by making use of these tabular data.

% tabular foundation model
The summary feature vectors are a tabular representation. Traditional approaches to tabular data have long been dominated by tree-based methods, such as gradient-boosted decision trees \citep{hollmann-iclr-tabpfn,hollmann-nature-tabpfn}. While deep learning has revolutionised learning from text and images, it has struggled with tabular data due to heterogeneity across datasets and the heterogeneity of raw data itself. Foundation models, such as \acp{LLM} and \acp{LVM}, have been widely explored for downstream tasks on text and video data. Similarly, \acp{LLM} have also been explored with tabular data \citep{pmlr-v206-hegselmann23a,fang2024llm-tabular}, but these approaches convert table rows into textual formats such as sentences. As \acp{LLM} do not natively support numerical tabular data, a new paradigm of the foundation model has emerged: the \acp{TFM}\footnote{Not to be confused with \acp{LLM}. While both are pre-trained \emph{foundation} models, \acp{TFM} are trained on numerical tabular data, whereas \acp{LLM} are trained on text. \acp{TFM} are significantly smaller than \acp{LLM}. For example, TabPFN has only 7.2 million trainable parameters, in contrast to billions in typical \acp{LLM}.}. The first of its kind, a \ac{PFN} \citep{muller2022iclr}, trained on extensive synthetic tabular data, is TabPFN \citep{hollmann-iclr-tabpfn}. Following the success of TabPFN, TabICL (Tabular \acl{ICL}) \citep{qu-arxiv-tabicl} is another \ac{TFM} that was introduced in 2025. 
% if space constraints, this can be moved to method.
In \acp{TFM}, \ac{ICL} refers to conditioning on a labelled set of input-output examples at inference time to produce predictions for new samples, which can be interpreted as amortised Bayesian-style inference \citep{muller2025position}. With these examples, the model adapts to new tasks without any gradient updates (\ie{} training). This differs from \ac{LLM} \ac{ICL}, where demonstrations are embedded in a text prompt to elicit a behaviour.

% brief of prior work and our contributions
\paragraph*{Novelty and Contributions}
Previous work on video-based elicited empathy detection demonstrated XGBoost, a classical \ac{ML} model, as the \ac{SOTA} method. In contrast, our proposed \acs{TFMPathy} system explores the potential of \acp{TFM} on this task and outperforms existing methods by a clear margin while still maintaining privacy (\Cref{subsec:res-kfold} and \Cref{subsec:res-generalisation}).
While \acp{TFM} are designed for numerical tabular data, we extend them to visual features, in a similar spirit to many works on using \acp{LLM} on new domains \citep{yang2022empirical,keesing2023emotion}.
We reframe video-based empathy detection under a three-level privacy taxonomy and show that strong, fair and generalisable empathy detection is achievable at the most restrictive (strong privacy) level. 
% Our final model, TabPFN fine-tuning, provides both strong performance and strong privacy for empathy detection in video interactions. 
To make fine-tuning effective and practical, we address key challenges such as catastrophic forgetting and conduct extensive ablation studies covering layer freezing, optimiser and scheduler choices, batch size, softmax temperature and feature dimensionality (\Cref{subsec:ablation}). We provide new insights and practical strategies for adapting \acp{TFM} to privacy-constrained, real-world video data. On the public \ac{HRI} dataset, we establish a cross-subject generalisation benchmark, which has not been evaluated before.
Our proposed fine-tuning of the \ac{TFM} achieves strong cross-subject generalisation (\Cref{subsec:res-generalisation}). 
Our key contributions are threefold:
\begin{enumerate}
    \item We formalise a three-level privacy taxonomy (\emph{no}, \emph{partial} and \emph{strong} privacy) for video-based behavioural prediction and explore \ac{ICL} of two recent \acp{TFM} (TabPFN and TabICL) on the strong-privacy representation for empathy detection. 
    \item We propose the \acs{TFMPathy} framework, which introduces a practical and effective fine-tuning strategy for TabPFN for empathy detection, achieving \ac{SOTA} performance on a public HRI benchmark while operating entirely on privacy-preserved summary features.
    \item We enhance the evaluation protocol of this benchmark by introducing subject-independent evaluation, which quantifies performance consistency across individuals -- a prerequisite for both individual-level fairness and safe deployment for real-world applications.
\end{enumerate}

\section{Related Work}

\subsection{Privacy and Fairness in Video-Based Affective Computing}
A growing body of work has examined privacy and fairness in affective computing pipelines. On the privacy side, the closest precursor to our setting is \citet{mase2023facial}, who compare two privacy-preserving strategies for valence/arousal prediction: extracting action units locally and discarding the raw images, and federated learning with locally trained model updates. Both strategies operate at what we term the partial-privacy level (frame-level features), and neither addresses the residual leakage of identity, demographics or health-related attributes that those features retain.

On the fairness side, \citet{xu2020investigating} proposed a bias mitigation strategy for facial expression recognition through supervised disentanglement of expression and demographic attributes.
Many benchmark datasets, including the \ac{HRI} dataset \citep{mathur-acii-hri}, release neither raw video nor demographic annotations, which motivates our particular approach to ensuring privacy and fairness using only available feature data. Prior work on the \ac{HRI} benchmark \citep{mathur-acii-hri} has reported only subject-unaware cross-validation and has not characterised either the privacy or the bias profile of the released representation. \ac{TFMPathy} addresses both concerns by learning from summary statistics (strong privacy) and evaluating cross-subject generalisation as a proxy for performance consistency across individuals.

\subsection{Empathy Detection}
Several empathy detection benchmarks are proposed in the literature, aiming at different aspects of empathy such as assessing the level of empathy in an interaction \citep{buechel2018modeling,mathur-acii-hri}, measuring empathic similarity between two contents \citep{shen-etal-2023-modeling}, measuring the changes of someone's emotional state \citep{barros2019omg}, and detecting the direction of empathy \citep{hosseini2021it}. Our work is grounded on the assessment of the \emph{level} of empathy, which is the most common form in the literature \citep{hasan2025empathy}.

Several studies on empathy detection in text domains exist, including written essays \citep{buechel2018modeling,hasan2023curtin}, social media conversations \citep{khanpour2017identifying}, and email correspondence \citep{sedefoglu2024leadempathy}. In detecting empathy from videos, \citet{mathur-acii-hri} released the \iac{HRI} dataset to measure empathy elicited by a robot storyteller. They evaluated 10 different \ac{ML} approaches, including eight classical models (\eg{} AdaBoost, bagging, decision trees, support vector machine, logistic regression, random forest and XGBoost) and two deep learning architectures (\ac{LSTM} and \ac{TCN}). For classical \ac{ML} approaches, features were represented as fixed-length vectors of statistical attributes (mean, median, standard deviation and autocorrelation). For deep learning approaches, the features were maintained as temporal sequences. In 5-fold stratified cross-validation with 10 repeats, XGBoost performed best. Contrary to their work, we propose \ac{ICL} and fine-tuning of \acp{TFM} in detecting elicited empathy. 
% They first extracted 709 visual features and then selected 25 features, which were inputted into the models.

Other studies on video-based empathy detection include interaction between human subjects and virtual agents \citep{kroes2022empathizing,tavabi2019multimodal}, teacher and students \citep{pan2022multimodal}, human subject and avatar \citep{hervas2016learning}, counsellor and clients \citep{zhu2023medic}, two human subjects \citep{barros2019omg}, and human and AI agents \citep{shen-etal-2024-empathicstories}. In contrast to these empathy-modelling approaches, we focus on measuring elicited empathy from the public in the \ac{HRI} benchmark.

\subsection{Application of Foundation Models}
Textual foundation models (\ie{} \acp{LLM}) have been explored in several stages of empathy detection from textual data. For example, our prior work \citep{hasan2026labels} used \acr{GPT}-4 and Llama-3 \acp{LLM}, and \citet{kong-moon-2024-ru} used \acs{GPT} series \acp{LLM} in a zero-shot setting to predict empathy in written essays. On the same task, \citet{kong-moon-2024-ru} and \citet{hasan2024llm-gem} explored few-shot prompt engineering (\ie{} \ac{ICL}) with \acs{GPT} \acp{LLM}. Apart from the prediction, \acp{LLM} are also explored in data augmentation \citep{lu2023hit,frick-steinebach-2024-fraunhofer} and data annotation \citep{hasan2026labels,hasan2024llm-gem}. 

Motivated by the success of \acp{LLM} in empathy detection from text, we explore two recent \acp{TFM} released in 2025 -- TabPFN v2 \citep{hollmann-nature-tabpfn} and TabICL \citep{qu-arxiv-tabicl} -- for tabular video features. While \acp{TFM}' \ac{ICL} capacity has been demonstrated in numerical datasets such as OpenML test suites \citep{openml2014}, we explore its capability in dealing with video features. In addition to \ac{ICL}, we fine-tune TabPFN v2, which further boosts empathy classification performance.
 % \citep{hollmann-nature-tabpfn} introduced the Tabular Prior-data Fitted Network (TabPFN), a foundation model for tabular data that leverages \ac{ICL} (ICL) to make predictions without requiring explicit training on new datasets. In their experiment, TabPFN outperforms traditional methods on datasets with up to 10,000 samples and 500 features, achieving better results than gradient-boosted decision trees in substantially less training time.

\section{Method}

\begin{figure*}[t!]
    \centering
    \includegraphics[trim=1.8cm 0 1.2cm 0, clip, width=0.9\linewidth]{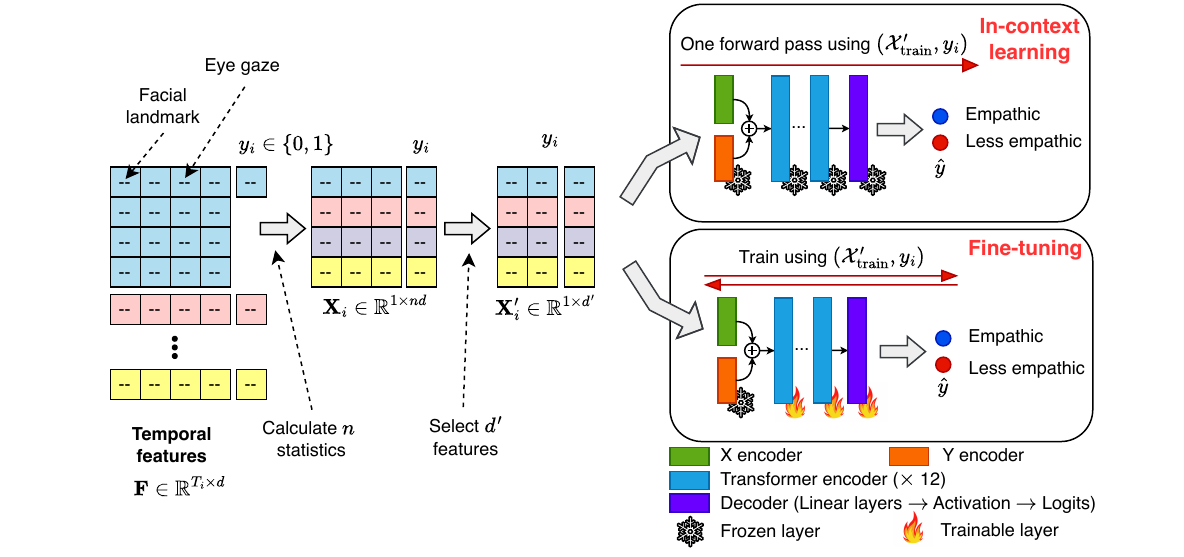}
    \caption{Overview of our proposed \acs{TFMPathy} pipeline for strong privacy-preserving empathy detection. Temporal multimodal features $\mathbf{F}_i \in \mathbb{R}^{T_i \times d}$ (\eg{} facial landmarks, eye gaze) are first extracted for each sample. These sequences are summarised by computing $n$ statistical attributes (\eg{} mean, median) over time, resulting in a fixed-size tabular representation $\mathbf{X}_i \in \mathbb{R}^{1 \times nd}$. Following feature selection, the final representation $X'$ is used for binary empathy classification $y_i \in \{0,1\}$) via either (top) \ac{ICL}, where a frozen \ac{TFM} predicts labels with a single forward pass over the training data, or (bottom) fine-tuning of the foundation model with frozen X and Y encoders. The model predicts whether a subject is empathic or less empathic.}
    \label{fig:overview}
\end{figure*}

\Cref{fig:overview} presents an overview of our approach for empathy detection using \acp{TFM}. Temporal multimodal features, such as facial landmarks and eye gaze, are first aggregated into fixed-size tabular representations via summary statistics. These are then used for binary empathy classification through either \ac{ICL} or fine-tuning of \iac{TFM}. Further details of each component are provided in the following subsections.

\subsection{Privacy Modelling \& Problem Formulation}
We adopt a practical, data-minimisation view of privacy rather than a formal cryptographic or differentially-private guarantee. Concretely, we consider an adversary who has access to the released dataset and, possibly to the trained classifier, and seeks either to re-identify individual subjects or to reconstruct their visual appearance. Under this view, we distinguish three nested regimes of increasing privacy, illustrated in \Cref{fig:overview}.

\paragraph{No privacy (raw video)} The representation is a sequence of raw pixel frames $V\in\mathbb{R}^{T\times H\times W\times 3}$. This preserves facial appearance, expressions, background and dynamics within the biometric data categories treated as sensitive under the GDPR \citep{it2020eu}.

\paragraph{Partial privacy (temporal visual features)} The representation is a sequence of frame-level visual features $F\in\mathbb{R}^{T\times d}$ -- eye gaze, facial action units, facial landmarks and head pose -- extracted by a toolkit such as OpenFace \citep{amos2016openface}. Raw pixels are discarded, but per-frame geometry, orientation and dynamics are retained. These retained signals could be used for face reconstruction from landmarks \citep{di2018gpgan}, for biometric identification and demographic inference from gaze patterns \citep{kroger2020what}, and more generally for subject re-identification in affective computing pipelines \cite{mase2023facial}. We therefore treat
this level as \emph{partial} privacy-preserving.

\paragraph{Strong privacy (summary statistics)} The representation is a fixed-size vector of statistical attributes aggregated across the temporal axis: $X\in\mathbb{R}^{1\times nd}$ (\eg{} mean, median, standard deviation, autocorrelation). Aggregation is lossy and non-invertible: the fine-grained temporal trajectories on which per-subject biometrics rely are collapsed into summaries. This substantially reduces the residual re-identification and demographic inference risks, but does not eliminate them.

In terms of information content, these regimes form an approximately decreasing-information hierarchy $V\supseteq F\supseteq X$: any utility achievable on a lower-information representation provides a lower bound on the utility achievable with any higher-information representation, but at the cost of greater privacy risk. The central empirical question of this paper is how much classification utility TFMPathy can recover at the strong-privacy level.

Having defined the three privacy regimes, we now formulate empathy detection at the strong-privacy level as a supervised or an \ac{ICL} problem. Given a dataset $\mathcal{D} = \{(\mathbf{F}_i, y_i)\}_{i=1}^N$, consisting of $N$ interaction samples from $M$ subjects, where $\mathbf{F}_i \in \mathbb{R}^{T_i \times d}$ represents a $d$-dimensional feature vector across $T_i$ frames of the $i$-th video sample and $y_i \in \{0, 1\}$ denotes the corresponding binary empathy label.

\subsection{Data Preprocessing for Tabular Foundation Model}
The feature vector $\mathbf{F}_i$ contains frame-level visual behavioural cues (\eg{} eye gaze directions, facial action unit intensities, facial landmarks, and other non-verbal indicators) extracted using toolkits such as OpenFace \citep{amos2016openface}. These temporal features can be processed in two ways: (1) time-series modelling or (2) transformed into fixed-length representations by computing statistical attributes (\eg{} mean, median, standard deviation and autocorrelation) across the temporal axis. The prior work \citet{mathur-acii-hri}, which released the \ac{HRI} dataset, employed time-series modelling with \ac{LSTM} and \ac{TCN} models, as well as fixed-length statistical summaries in classical models. Their experiments showed better performance with the latter approach, so we also adopt it.

From $\mathbf{F}_i \in \mathbb{R}^{T_i \times d}$, we first filter out temporal points having zeros in all features, and then calculate statistical attributes. For $n$ statistical attributes, the input feature representation becomes a fixed-size representation $\mathbf{X}_i \in \mathbb{R}^{1\times nd}$. For reference, in OpenFace features (709 dimensions) and four statistical attributes, the resultant feature becomes a vector of $709\times4=2836$ elements. From this, we first drop dimensions that are constant across all samples.

Both TabPFN and TabICL support up to 500 features, and therefore, we further reduce the feature dimension. We rank all features by their ANOVA F‑values using the target labels from the training set only, and retain the top $d'$ features ($X'$). At evaluation time, we apply this fixed subset to each test split, without recomputing F‑values to prevent any information leakage from the test to the training set. This feature selection process is performed independently within each cross-validation fold to avoid data leakage.

With this tabular representation, our goal is to learn a function $f: \mathbb{R}^{1\times d'} \rightarrow [0, 1]$ that maps the input features to a conditional probability distribution over the target variable (empathy level). For a general supervised learning task, it becomes:
\begin{equation}
p(y=1|\mathbf{x}; \theta) = f_\theta(\mathbf{x})
\end{equation}
where $\mathbf{x} \in \mathbf{X'}$ represents the input features, and $\theta$ denotes the model parameters optimised during training.

\subsection{Traditional Supervised Learning Approaches}
In supervised learning, we use labelled training data $\mathcal{D}_{\text{train}}$ to optimise model parameters:
\begin{equation}
\theta^* = \arg\min_\theta \mathcal{L}(\theta, \mathcal{D}_{\text{train}})
\end{equation}
where $\mathcal{L}$ is a loss function (\eg{} binary cross-entropy) and $\theta^*$ represents the optimised parameters. During inference, predictions are made on a separate test set:
\begin{equation}
p(y=1|\mathbf{x}; \theta^*) = f_{\theta^*}(\mathbf{x})
\end{equation}

\subsection{In-Context Learning with Tabular Foundation Models}
\acp{TFM} such as TabPFN and TabICL are pre-trained on large collections of synthetic tabular tasks and, at inference time, predict labels for query samples by conditioning on a labelled support set as context (we use the training set $\mathcal{D}_{\text{train}}$ as the context). This approach can be viewed as amortised Bayesian posterior predictive inference under an implicit prior over tabular tasks \citep{muller2025position}\footnote{Note that, although \ac{ICL} also exists in \acp{LLM}, \ac{LLM} \ac{ICL} involves conditioning on natural-language demonstrations in a prompt to elicit a behaviour, rather than being framed as posterior inference over a supervised dataset.}. Formally, they approximate the posterior predictive distribution:
\begin{equation}
p(y_{\text{test}}|\mathbf{x}_{\text{test}}, \mathcal{D}_{\text{train}}) \approx q_\theta(y_{\text{test}}|\mathbf{x}_{\text{test}}, \mathcal{D}_{\text{train}})
\end{equation}
where $q_\theta$ is parameterised by a transformer-based neural network with parameters $\theta$. For binary empathy classification, the prediction from TabPFN is:
\begin{equation}
\hat{y} = \arg\max_{y \in \{0,1\}} q_\theta(y|\mathbf{x}_{\text{test}}, \mathcal{D}_{\text{train}})
\end{equation}

A simplified representation of the TabPFN architecture is included in \Cref{fig:overview} (refer to \citep{hollmann-iclr-tabpfn, hollmann-nature-tabpfn} for details). The model includes two separate encoders (X and Y) for processing features and labels, respectively, which also perform basic preprocessing such as data cleaning and input normalisation. The encoded features and labels are concatenated and passed through 12 transformer blocks. Each block includes two multi-head attention mechanisms: one for feature-wise attention and another for item-wise attention. Finally, a decoder transforms the output using a linear layer followed by GELU activation.

\subsection{Fine-Tuning Tabular Foundation Model}
While \acp{TFM} achieve strong \ac{ICL} performance, it can be further enhanced by fine-tuning on the training data. We update the model parameters $\theta$ to $\theta'$ by minimising the following binary cross-entropy loss with sigmoid activation and temperature scaling:
\begin{multline}
\mathcal{L}_{\text{fine}} = -\frac{1}{N_{\text{fine}}}\sum_{i=1}^{N_{\text{fine}}} \Bigl[ y_i \cdot \log(\sigma(z_i^+/\tau)) \\  + (1 - y_i) \cdot \log(1 - \sigma(z_i^+/\tau)) \Bigr]
\end{multline}
where $z_i^+$ represents the positive class logit for the $i$-th sample, $\sigma(z) = \frac{1}{1 + e^{-z}}$ is the sigmoid function, $N_\text{fine}$ is the number of samples in fine-tuning, and $\tau$ controls the smoothness of the probability distribution.

Catastrophic forgetting is a common challenge with fine-tuning pre-trained models. To address this, we apply a warm-up strategy (where the learning rate is linearly increased from a small initial value) with the optimiser. Specifically, we used a one-cycle learning rate scheduler with cosine annealing. To prevent overfitting during fine-tuning, we employ adaptive early stopping, which dynamically adjusts the patience parameter (\ie{} the number of steps before stopping the training) throughout training rather than relying on a fixed value. We save the best model based on validation performance.

\subsection{Evaluation Protocol}
We employ two complementary cross-validation approaches: k-fold stratified cross-validation and \ac{LOSO} cross-validation. In each approach, we calculate five evaluation metrics: classification accuracy, precision, recall, F1 score, and \ac{AUC}.

% \subsubsection{k-fold stratified cross-validation}
K-fold stratified cross-validation is shown to be a suitable technique for evaluating models when the sample size is small. Although stratified sampling maintains the class distribution of the original dataset across all folds, it has limitations when applied to datasets with multiple samples per subject. Specifically, samples from the same subject may appear in both training and validation sets, potentially leading to an overly optimistic evaluation.
% This occurs because:
% \begin{equation}
% \begin{split}
% \exists (\mathbf{x}_i, y_i), & (\mathbf{x}_j, y_j) \in \mathcal{D}: \\
% \text{subject}(\mathbf{x}_i) = & \text{subject}(\mathbf{x}_j) \land (\mathbf{x}_i, y_i) \in \mathcal{D}_{\text{train}}^{(f)} \land (\mathbf{x}_j, y_j) \in \mathcal{D}_{\text{test}}^{(f)}
% \end{split}
% \end{equation}

% \subsubsection{Leave-One-subject-Out Cross-Validation}
To address this limitation and evaluate the model's ability to generalise to unseen individuals, we adopt \ac{LOSO} cross-validation. In this setting, each fold corresponds to a held-out person; variability across folds therefore reflects how unevenly the model's errors are distributed across individuals. We interpret cross-subject generalisation not only as robustness to subject shift, but as a prerequisite for fair and safe deployment in diverse populations, where the system must perform reliably for previously unseen users.
Let $\mathcal{S} = \{s_1, s_2, \ldots, s_M\}$ represent the set of $M$ subjects, and $\mathcal{D}_{s_j}$ denote the subset of samples from subject $s_j$. For each subject $s_j$, we define:
\begin{equation}
\mathcal{D}_{\text{train}}^{(j)} = \mathcal{D} \setminus \mathcal{D}_{s_j}, \quad \mathcal{D}_{\text{test}}^{(j)} = \mathcal{D}_{s_j}
\end{equation}

This formulation ensures complete separation of subjects between training and test sets:
\begin{equation}
\forall (\mathbf{x}_i, y_i) \in \mathcal{D}_{\text{train}}^{(j)}, (\mathbf{x}_j, y_j) \in \mathcal{D}_{\text{test}}^{(j)}: \text{subject}(\mathbf{x}_i) \neq \text{subject}(\mathbf{x}_j)
\end{equation}

As stratification cannot be ensured in this cross-validation, some test splits may have all samples from the same class. This is more likely when subjects have a small number of samples. Some evaluation metrics, such as \ac{AUC}, are undefined in this scenario. To address this edge case, we pool predictions across all subject folds to calculate the evaluation metrics.

\section{Experiments and Results}
\subsection{Dataset}
We use the \ac{HRI} dataset \citep{mathur-acii-hri}. In each interaction, a participant listened to three different stories, each lasting approximately 3 minutes. 
Having 4 videos removed due to technical issues, the final dataset includes 122 interaction samples (a total of 6.9 hours) between 42 human participants and a robot storyteller.
The stories were delivered under randomly assigned narrative voice conditions: half in a first-person narrative voice, in which the robot told stories about itself, and half in a third-person narrative voice, in which the robot narrated from an external perspective. After each story, participants completed an 8-item Likert-scale questionnaire, yielding a ground-truth empathy score for each interaction. \citet{mathur-acii-hri} binarised the labels into ``empathic'' and ``less-empathic'' categories using the median empathy score as a threshold. The label distribution is balanced (61 empathic and 61 less-empathic samples).

The dataset is released publicly as extracted features. For each video, 709 visual behavioural features were extracted using the OpenFace 2.2.0 toolkit \citep{amos2016openface}, which captured comprehensive facial and gaze information from human subjects as they listened to the robot's stories. Features include eye gaze direction vectors (per eye); facial action unit intensity and presence; facial landmarks; head pose rotation and translation; and point distribution model parameters representing facial location, scale, rotation, and non-rigid deformation.

\begin{table}[!t]
    \centering
    \caption{Hyperparameter tuning results of 50 trials in Optuna, optimised for 5-fold 10-repeats stratified cross-validated average classification accuracy. \textbf{Bold} represents hyperparameters corresponding to the best accuracy.}
    \label{tab-supp:hparams}
    % \resizebox{0.8\linewidth}{!}{%
    \begin{tabular}{@{}*2lc@{}} \toprule
        \textbf{Model} & \textbf{Hyperparameter} & \textbf{Ranges of values} \\ \midrule
        XGBoost & Max depth & \{3, 4, ..., \textbf{42}, ..., 50\} \\
            & Learning rate & [0.01, ..., \textbf{0.046}, ..., 0.5] \\ \midrule
        RF & Number of estimators & \{50, 100, \textbf{150}, ..., 500\} \\
            & Max depth & \{3, 4, ..., \textbf{34}, ..., 50\} \\ \midrule
        SVM & C & [0.1, ..., \textbf{1.0}, ..., 100] (log scale) \\
            & Gamma & \textbf{`scale'}, [0.001, ..., 10] (log scale) \\ \midrule
        TabPFN & Learning rate & [1e-6, ..., \textbf{1e-4}, ..., 1e-2] (log scale) \\
            & Batch size & \{8, 16, 24, \textbf{32}, ..., 64] \\\bottomrule
    \end{tabular}%
    % }
\end{table}

\begin{table*}[t!]
    \centering
    \caption{5-fold 10-repeat stratified cross-validation (average $\pm$ standard deviation) scores of our proposed \acs{TFMPathy} versus baseline models. Single scores refer to the scores reported in the literature. \textbf{Bold} indicates the highest scores in each configuration.}
    \label{tab:res-kfold}
    % \resizebox{0.8\textwidth}{!}{%
    \begin{threeparttable}
    \begin{tblr}{
        colspec={X[l] *{5}{X[c]}},
        % colspec={Q[l] *{5}{Q[c]}},
        % width=\textwidth
    }\toprule
        \textbf{Model} & \textbf{Accuracy} & \textbf{\acs*{AUC}} & \textbf{Precision} & \textbf{Recall} & \textbf{F1 Score} \\ \midrule
        \SetCell[c=6]{l}\textit{\textbf{Partial privacy} (learning from temporal features)} \\
        \acs*{LSTM} \citep{mathur-acii-hri} & \textbf{0.65} & \textbf{0.71} & 0.65 & \textbf{0.70} & 0.67 \\
        \acs{LSTM}\tnote{a} & $0.623 \pm 0.103$ & $0.666 \pm 0.113$ & $0.636 \pm 0.123$ & $0.606 \pm 0.164$ & $0.609 \pm 0.122$ \\
        \acs*{TCN} \citep{mathur-acii-hri} & 0.54 & 0.54 & \textbf{0.76} & 0.62 & \textbf{0.68} \\
        \acs{TCN}\tnote{a} & $0.613 \pm 0.098$ & $0.659 \pm 0.105$ & $0.611 \pm 0.101$ & $0.666 \pm 0.123$ & $0.631 \pm 0.092$ \\
        Transformer & $0.631 \pm 0.096$ & $0.679 \pm 0.108$ & $0.631 \pm 0.103$ & $0.678 \pm 0.144$ & $0.644 \pm 0.094$ \\
        
        \midrule
        \SetCell[c=6]{l}\textit{\textbf{Strong privacy} (learning from summary features)} \\
        XGBoost \citep{mathur-acii-hri} & 0.69 & \textbf{0.72} & 0.69 & \textbf{0.69} & 0.69 \\
        XGBoost \citep{mathur-acii-hri}\tnote{a} & $0.598 \pm 0.071$ & $0.623 \pm 0.099$ & $0.604 \pm 0.079$ & $0.589 \pm 0.135$ & $0.588 \pm 0.090$ \\
        % XGBoost\tnote{c} & $0.663 \pm 0.111$ & $0.705 \pm 0.131$ & $0.659 \pm 0.114$ & $\textbf{0.701} \pm 0.210$ & $0.663 \pm 0.136$ \\
        XGBoost\tnote{b} & $0.612 \pm 0.093$ & $0.651 \pm 0.107$ & $0.611 \pm 0.098$ & $0.638 \pm 0.161$ & $0.615 \pm 0.107$ \\
        \acs{RF} & $0.625 \pm 0.109$ & $0.651 \pm 0.115$ & $0.631 \pm 0.119$ & $0.639 \pm 0.135$ & $0.628 \pm 0.110$ \\
        \acs{SVM} & $0.605 \pm 0.080$ & $0.641 \pm 0.098$ & $0.619 \pm 0.099$ & $0.581 \pm 0.119$ & $0.592 \pm 0.089$ \\
        \acs{TFMPathy}\textsubscript{ICL}(TabICL) & $0.634 \pm 0.084$ & $0.665 \pm 0.103$ & $0.637 \pm 0.100$ & $0.660 \pm 0.139$ & $0.639 \pm 0.094$ \\
        \acs{TFMPathy}\textsubscript{ICL}(TabPFN) & $0.622 \pm 0.091$ & $0.653 \pm 0.100$ & $0.621 \pm 0.096$ & $0.651 \pm 0.148$ & $0.627 \pm 0.103$  \\ 
        \acs{TFMPathy}\textsubscript{FT}(TabPFN) & $\textbf{0.715} \pm 0.073$ & $0.700 \pm 0.091$ & $\textbf{0.773} \pm 0.126$ & $0.659 \pm 0.156$ & $\textbf{0.691} \pm 0.096$ \\
        \bottomrule
    \end{tblr}
    \begin{tablenotes}
        \item Single scores represent scores reported in literature without any specification of metric type (average or best).
        \item \acs{ICL} -- \acl{ICL}, FT -- Fine-Tuning.
        \item[a] Our re-implementation of the earlier method \citep{mathur-acii-hri} based on their limited available hyperparameters (code or full hyperparameter details were unavailable).
        \item[b] Our implementation of the earlier SOTA method with our own hyperparameter optimisation.
    \end{tablenotes}
    \end{threeparttable}%
    % }
\end{table*}

% \begin{table*}[t!]
%     \centering
%     \caption{MEDIC - 5-fold 10-repeat stratified cross-validation (average $\pm$ standard deviation) scores of our proposed \acs{TFMPathy} versus baseline models. Single scores refer to the scores reported in the literature. Highest scores are formatted in \textbf{bold}. \ac{ICL} refers to in-context learning, and FT refers to fine-tuning.}
%     \label{tab:res-kfold}
%     \begin{threeparttable}
%     \begin{tabular}{*1l*5c} \toprule
%         \textbf{Model} & \textbf{Accuracy} & \textbf{\acs*{AUC}} & \textbf{Precision} & \textbf{Recall} & \textbf{F1 Score} \\ \midrule
%         \textit{Classical models} \\

%         XGBoost & $0.612 \pm 0.093$ & $0.651 \pm 0.107$ & $0.611 \pm 0.098$ & $0.638 \pm 0.161$ & $0.615 \pm 0.107$ \\
%         RF &  \\
%         SVM &  \\
%         \midrule
%         \textit{Deep learning} \\
%         \ac*{LSTM} \citep{mathur-acii-hri} \\
%         \ac*{TCN} \\
%         \acs{TFMPathy}\textsubscript{ICL}(TabICL) & $0.645 \pm 0.034$ & $0.667 \pm 0.043$ & $0.595 \pm 0.076$ & $0.376 \pm 0.050$ & $0.458 \pm 0.049$ \\
%         \acs{TFMPathy}\textsubscript{ICL}(TabPFN) & $0.637 \pm 0.034$ & $0.649 \pm 0.043$ & $0.582 \pm 0.078$ & $0.351 \pm 0.071$ & $0.434 \pm 0.066$ \\ 
%         \acs{TFMPathy}\textsubscript{FT}(TabPFN) & $0.645 \pm 0.023$ & $0.611 \pm 0.042$ & $0.661 \pm 0.115$ & $0.312 \pm 0.146$ & $0.390 \pm 0.136$ \\
%         \bottomrule
%     \end{tabular}
%     \begin{tablenotes}
%         \item[a] 
%     \end{tablenotes}
%     \end{threeparttable}
% \end{table*}

\subsection{Selection of Baseline Models}
The \textit{partial privacy} setup takes temporal sequence as input, so we select sequence learning baselines models, including \ac{LSTM}, \ac{TCN} and a Transformer network.
While advanced video models could, in principle, capture richer temporal cues, they would require access to the full video sequences and therefore not privacy-aware.
Our privacy-preserving framework also aims to address the unavailability of raw videos due to privacy and ethics constraints. Specifically for the \ac{HRI} benchmark \citep{mathur-acii-hri}, raw videos are not released as a data minimisation measure to reduce privacy risk, and tabular representations of pre-extracted visual features are the only available modality. All models experimented in this paper are therefore designed to work with the available visual features but not any raw videos.

The \textit{strong privacy} setup takes summary features to ensure privacy-preserving learning and inference. Such summary features resemble the characteristics of tabular data. In tabular datasets, gradient boosting methods continue to dominate \citep{hollmann-iclr-tabpfn}. The original work proposing TabPFN \citep{hollmann-iclr-tabpfn} benchmarks TabPFN against many gradient boosting methods, including XGBoost, across 18 numerical datasets.
We therefore benchmark our \ac{TFM}-based privacy-preserving framework against XGBoost, \ac{RF} and \ac{SVM}.
% Prior research \citep{mathur-acii-hri} on the \ac{HRI} benchmark also found that XGBoost was the best method, outperforming deep learning models. 
% Therefore, the above selection of baseline models is appropriate given the benchmark's feature-only nature.

\subsection{Implementation Details}
We use PyTorch 2.2.0, scikit-learn 1.6.1 and other relevant libraries under Python 3.10.12 to implement the empathy detection workflow and to run the experiments. Details of other libraries and their versions are provided in our public GitHub repository. Experiments are run on a single AMD Instinct™ MI250X GPU (64 GB GCD).

Hyperparameters are tuned using Optuna \citep{akiba2019optuna} with its default sampler (tree-structured Parzen estimator) and pruner (median stopping rule). For a fair comparison, we optimise two key hyperparameters across 50 trials for all baseline models and the TabPFN fine-tuning setup. The objective of the optimisation is to maximise the average classification accuracy across 5-fold, 10-repeats stratified cross-validation. We did not tune the hyperparameters of TabICL and TabPFN in the \ac{ICL} setup; instead, we used their default configurations from their respective official libraries. \Cref{tab-supp:hparams} presents the list of hyperparameters and corresponding models. The same set of selected hyperparameters is then applied in our \ac{LOSO} setup to experiment with generalisation to unseen subjects.

\subsubsection{Tabular Foundation Models} We use the official pre-trained release of TabPFN v2\footnote{\url{https://huggingface.co/Prior-Labs/TabPFN-v2-clf}} and TabICL\footnote{\url{https://huggingface.co/jingang/TabICL-clf}} foundation models. Since pre-trained TabICL models were unavailable at the time of this work, TabICL was used only in the \ac{ICL} experiment, whereas TabPFN was experimented with both \ac{ICL} and fine-tuning settings.

\subsubsection{Implementation of Baseline Models} We benchmark against several classical machine learning and deep learning models. 
The work releasing \ac{HRI} dataset \citep{mathur-acii-hri} reported XGBoost as the best model for this dataset. Since no official implementation of their XGBoost configuration was released, we re-implemented it using the three hyperparameters specified in \citep{mathur-acii-hri}. For other hyperparameters not specified in \citep{mathur-acii-hri}, we use default values from the official XGBoost library. 
% We denote this version as XGBoost\textsuperscript{\citep{mathur-acii-hri}} throughout this paper. 
We further implement an additional XGBoost model with hyperparameter optimisation via Optuna. Apart from XGBoost, we further implemented \ac{RF} and \ac{SVM} algorithms with their default hyperparameters as baseline models.

\subsection{Results: Better Performance on Established Evaluation Protocol While Ensuring Strong Privacy}\label{subsec:res-kfold}
Following the 5-fold 10-repeats stratified cross-validation evaluation protocol established in \citet{mathur-acii-hri}, \Cref{tab:res-kfold} compares cross-validated average and standard deviation for both partial-privacy and strong-privacy setups. Since TabPFN's recommended input capacity is up to 500 features \citep{hollmann-nature-tabpfn}, we select the 500 best features. For a fair comparison, we also select 500 features for strong-privacy baseline models, including XGBoost, \ac{RF} and \ac{SVM}. For the strong privacy setup, the scores of \ac{ICL} and fine-tuning of \acs{TFMPathy} are benchmarked against baseline models as well as the performance reported in the literature.

In partial privacy, LSTM achieved the highest accuracy of 0.65 and an AUC score of 0.71. These scores are surpassed by TFMPathy in fine-tuning setups while ensuring strong privacy. Our proposed TFMPathy therefore provides both better classification performance and better privacy. In terms of Precision and F1 scores, TFMPathy outperforms partial privacy modelling; however, the recall is slightly higher in partial privacy than in strong privacy, which resonates with the privacy-utility trade-off \citep{shaham2025privacy}.

The scores reported in the prior SOTA work \citep{mathur-acii-hri} do not specify whether it is the average or the highest score. Considering it is an average score across 50 folds, our fine-tuned TabPFN still outperforms it in classification accuracy, precision and F1 scores. 
The discrepancy between the reported scores for XGBoost \citep{mathur-acii-hri} and our re-implementation is likely due to undisclosed hyperparameters or training details.
% Nonetheless, the similarity between their reported results and our optimised XGBoost variant of XGBoost highlights the models' known strength with tabular data.

\textit{Fine-tuning tabular foundation model further improves performance.} When fine-tuned, TabPFN shows a substantial performance gain and surpasses TabICL. This improvement, however, comes at the cost of additional computation compared to the \ac{ICL} setup. TabPFN has 7.2 million trainable parameters, and in our experiment on a 64 GB GPU, the average fine‑tuning time per step ranged from 3.86 to 4.08 seconds for each of the 50 folds. However, compared to no-privacy, end-to-end \acp{LVM} pipelines that require processing full video sequences and adapting large spatiotemporal backbones, our approach remains comparatively lightweight because it operates on fixed, pre-extracted tabular features and adapts a compact tabular model. This also connects to Green AI: the results suggest that strong empathy detection performance can be achieved without the computational footprint typically associated with \acp{LVM} training. Therefore, our proposed TFMPathy offers multiple benefits: privacy preservation, lower computational cost than no-privacy models, and better performance than partial-privacy setups.

\textit{TabICL outperforms TabPFN.} Consistent with the findings of \citep{qu-arxiv-tabicl}, our experiment confirms the superior performance of TabICL over TabPFN across all five evaluation metrics. Due to the lack of any pre-trained TabICL checkpoints at the time of this work, it was not fine-tuned. Since our experiments suggest that fine-tuning improves performance, there is strong potential to further improve performance with fine-tuning TabICL.
% While these results are not with any hyperparameter tuning for the \ac{ICL} experiments. 

\textit{\acs{TFMPathy} outperforms other deep learning approaches.} As shown in \Cref{tab:res-kfold}, \acs{TFMPathy} in strong-privacy setup outperforms \ac{LSTM}, \ac{TCN} and Transformer deep learning architectures.
% Their lower performance may be attributed to using raw temporal sequences as input rather than statistical attributes.
Their lower performance in this 5-fold setting is consistent with sequence models being data-hungry: with only $\approx 98$ training samples per fold, they struggle to learn temporal patterns from scratch, whereas TabPFN's pre-training on synthetic tabular tasks provides a strong inductive prior that transfers effectively in this low-data regime.
% In LOSO (Section IV-E), where the per-fold training set is larger (~119 samples) and the protocol structurally rewards subject-agnostic patterns, the sequence models recover competitive performance — consistent with the privacy–utility trade-off

\begin{table}[t!]
    \centering
    \scriptsize
    \caption{Generalisation on unseen subjects through \ac{LOSO} cross-validation. Highest scores in each configuration are highlighted in \textbf{bold}.}
    \label{tab:result-logo}
    % \resizebox{1.0\textwidth}{!}{%
    \begin{threeparttable}
    \begin{tblr}{
        colspec={X[10, l] *{5}{X[c]}},
        % hline{1}={1pt},
        % width={0.65\textwidth}, % for full-width
        % width={0.65\linewidth},
        row{1}={font=\bfseries}
    } \toprule
        Model & Acc & \acs*{AUC} & P & R & F1 \\ \midrule
        \SetCell[c=6]{l}\textit{\textbf{Partial privacy} (learning from temporal features)} \\
        \acs{LSTM} & 0.648 & 0.787 & 0.680 & 0.557 & 0.613 \\
        \acs{TCN} & 0.705 & 0.788 & 0.705 & 0.705 & 0.705 \\
        Transformer & \textbf{0.746} & \textbf{0.827} & \textbf{0.742} & \textbf{0.754} & \textbf{0.748} \\
        \midrule
        
        \SetCell[c=6]{l}{\textit{\textbf{Strong privacy} (learning from summary features)}} \\
        XGBoost & 0.516 & 0.511 & 0.519 & 0.443 & 0.478 \\
        \acs{RF} & 0.533 & 0.523 & 0.536 & 0.492 & 0.513 \\
        \acs{SVM} & 0.590 & 0.564 & 0.590 & \textbf{0.590} & 0.590 \\
        \acs{TFMPathy}\textsubscript{\acs{ICL}}(TabICL) & 0.541 & 0.519 & 0.547 & 0.475 & 0.509 \\
        \acs{TFMPathy}\textsubscript{\acs{ICL}}(TabPFN) & 0.459 & 0.472 & 0.451 & 0.377 & 0.411 \\ 
        \acs{TFMPathy}\textsubscript{FT}(TabPFN; all layers trainable) & 0.689 & 0.642 & 0.780 & 0.525 & 0.627 \\
        \acs{TFMPathy}\textsubscript{FT}(TabPFN; frozen X and Y encoders) & \textbf{0.730} & \textbf{0.669} & \textbf{0.818} & \textbf{0.590} & \textbf{0.686} \\ \bottomrule
    \end{tblr}
    \begin{tablenotes}
        \item Acc -- Accuracy, P -- Precision, R -- Recall, F1 -- F1 Score
        \item \acs{ICL} -- \acl{ICL}, FT -- Fine-Tuning
    \end{tablenotes}
    \end{threeparttable}%
    % }
\end{table}

\begin{table*}[t!]
    \centering
    \caption{TabPFN fine-tuning at different experimental settings under strong privacy and subject-unaware 5-fold 10-repeats stratified cross-validation setup. \textbf{Bold} scores indicates the best scores and \textbf{Bold} configuration indicates the selected configuration.}
    \label{tab:tabpfn-ablation}
    % \resizebox{0.7\textwidth}{!}{%
    \begin{tabular}{*1l*5c} \toprule
        \textbf{Configuration} & \textbf{Accuracy} & \textbf{\acs*{AUC}} & \textbf{Precision} & \textbf{Recall} & \textbf{F1 Score} \\ \midrule
        \textit{Optimiser} \\
        Schedulerfree \citep{defazio2024road} & $\textbf{0.704} \pm 0.078$ & $0.685 \pm 0.102$ & $\textbf{0.770} \pm 0.122$ & $0.615 \pm 0.158$ & $0.666 \pm 0.112$ \\
        AdamW & $0.699 \pm 0.077$ & $0.688 \pm 0.100$ & $0.743 \pm 0.121$ & $\textbf{0.650} \pm 0.164$ & $\textbf{0.675} \pm 0.103$ \\
        \textbf{AdamW + OneCycleLR} & $0.701 \pm 0.077$ & $\textbf{0.689} \pm 0.095$ & $0.757 \pm 0.115$ & $0.624 \pm 0.148$ & $0.669 \pm 0.102$ \\ \midrule
        \textit{Batch size} \\
        8 & $0.701 \pm 0.077$ & $0.689 \pm 0.095$ & $\textbf{0.757} \pm 0.115$ & $0.624 \pm 0.148$ & $0.669 \pm 0.102$ \\
        \textbf{32} & $\textbf{0.708} \pm 0.078$ & $\textbf{0.696} \pm 0.091$ & $\textbf{0.757} \pm 0.116$ & $\textbf{0.656} \pm 0.178$ & $\textbf{0.680} \pm 0.118$ \\ \midrule
        \textit{Frozen layers (Batch size: 8)} \\
        None & $0.691 \pm 0.078$ & $0.665 \pm 0.108$ & $0.743 \pm 0.118$ & $0.623 \pm 0.179$ & $0.657 \pm 0.120$ \\
        X encoder, Y encoder, Transformer encoder & $0.639 \pm 0.084$ & $0.654 \pm 0.102$ & $0.675 \pm 0.133$ & $0.595 \pm 0.159$ & $0.615 \pm 0.104$ \\
        X encoder & $0.683 \pm 0.079$ & $0.661 \pm 0.107$ & $0.741 \pm 0.117$ & $0.602 \pm 0.190$ & $0.640 \pm 0.126$ \\
        Y encoder & $0.692 \pm 0.071$ & $0.672 \pm 0.100$ & $0.742 \pm 0.111$ & $\textbf{0.620} \pm 0.157$ & $\textbf{0.659} \pm 0.103$ \\
        Transformer encoder & $0.643 \pm 0.086$ & $0.654 \pm 0.102$ & $0.678 \pm 0.132$ & $0.599 \pm 0.164$ & $0.618 \pm 0.107$ \\
        \textbf{X encoder, Y encoder} & $\textbf{0.694} \pm 0.074$ & $\textbf{0.674} \pm 0.099$ & $\textbf{0.751} \pm 0.116$ & $0.610 \pm 0.159$ & $0.657 \pm 0.104$ \\ \midrule
        \textit{Softmax temperature (Batch size: 8)} \\
        \textbf{1.0} & $\textbf{0.691} \pm 0.078$ & $0.665 \pm 0.108$ & $0.743 \pm 0.118$ & $\textbf{0.623} \pm 0.179$ & $\textbf{0.657} \pm 0.120$ \\
        0.9 & $0.683 \pm 0.082$ & $0.667 \pm 0.110$ & $\textbf{0.748} \pm 0.123$ & $0.591 \pm 0.178$ & $0.638 \pm 0.125$ \\
        1.1 & $0.690 \pm 0.080$ & $\textbf{0.669} \pm 0.106$ & $0.735 \pm 0.123$ & $\textbf{0.623} \pm 0.182$ & $0.655 \pm 0.122$ \\
        1.5 & $0.683 \pm 0.072$ & $0.665 \pm 0.107$ & $0.730 \pm 0.116$ & $0.615 \pm 0.179$ & $0.646 \pm 0.122$ \\ \midrule
        \textit{Number of features selected (Batch size: 8)} \\
        25 & $0.691 \pm 0.078$ & $0.665 \pm 0.108$ & $0.743 \pm 0.118$ & $0.623 \pm 0.179$ & $0.657 \pm 0.120$ \\
        100 & $0.680 \pm 0.065$ & $0.669 \pm 0.089$ & $0.741 \pm 0.108$ & $0.593 \pm 0.171$ & $0.637 \pm 0.108$ \\
        200 & $0.691 \pm 0.059$ & $0.688 \pm 0.090$ & $0.731 \pm 0.104$ & $\textbf{0.648} \pm 0.141$ & $\textbf{0.671} \pm 0.076$ \\
        \textbf{500} & $\textbf{0.701} \pm 0.077$ & $\textbf{0.689} \pm 0.095$ & $\textbf{0.757} \pm 0.115$ & $0.624 \pm 0.148$ & $0.669 \pm 0.102$ \\
        % All & \\
        % \midrule
        % \textit{Mixed} \\ 
        % \textbf{Temperature 1, Batch size 32} & $0.708 \pm 0.078$ & $0.696 \pm 0.091$ & $0.757 \pm 0.116$ & $0.656 \pm 0.178$ & $0.680 \pm 0.118$ \\
        % Temperature 1.1, Batch size 32 & $0.707 \pm 0.083$ & $0.698 \pm 0.090$ & $0.755 \pm 0.122$ & $0.647 \pm 0.164$ & $0.679 \pm 0.123$ \\
    \bottomrule
    \end{tabular}%
    % }
\end{table*}

\subsection{Results: Improved Generalisation on Unseen Subjects}\label{subsec:res-generalisation}
Given the presence of multiple samples per subject (122 samples from 42 subjects), we adopt a train-test split that accounts for subject identity. Specifically, we use \ac{LOSO} cross-validation, where splitting is based on the participant ID available in the dataset. As this evaluation does not introduce any randomness, the results reported in \Cref{tab:result-logo} are deterministic and do not include standard deviation.

In partial-privacy experiments, the Transformer architecture achieved the best classification performance across all five evaluation metrics, outperforming \ac{LSTM} and \ac{TCN} by large margins. In the strong-privacy experiments, our proposed TFMPathy in the fine-tuning setting achieved the best performance across all five evaluation metrics.

\textit{Generalisation on unseen subjects is more challenging, yet more practical.} Models that achieve better performance under K-fold cross-validation -- without subject-specific separation -- struggled in this setting, which suggests the increased difficulty of the task. In this setting, models are tested on subjects whose data were absent during training, thereby avoiding inflated generalisation estimates. The evaluation metrics now better reflect the model's ability to learn subject-agnostic patterns, which is more realistic for practical deployment. It is also worth noting that, while \ac{ICL} does not involve supervised training, the model still ``sees'' the training data through the forward pass during the \textit{in-context} learning paradigm.

\textit{TabPFN in \ac{ICL} setup yields suboptimal performance.} While TabPFN in the \ac{ICL} setup performed competitively under subject-unaware (\ie{} K-fold) cross-validation, it performed poorly in the subject-aware (\ie{} \ac{LOSO}) cross-validation. With a baseline accuracy of 0.5 from a majority-class classifier (\ie{} a classifier that predicts all samples as either ``empathic'' or ``less empathic''), TabPFN \ac{ICL} achieved only 0.459 accuracy, which suggests it made more incorrect than correct predictions. Between TabPFN and TabICL, the latter performs better in the \ac{ICL} setting, consistent with the results from the subject-unaware cross-validation.

\textit{Fine-tuning gives a substantial boost to TabPFN's cross-subject generalisation.} Its classification accuracy improved from 0.459 to 0.730 after fine-tuning. Although the model had no access to the test subject's data, it effectively learned patterns from other subjects, suggesting good cross-subject generalisation. For each of the 42 subjects, the average fine‑tuning time per step ranged from 4.70 to 4.98 seconds, which is similar to the subject-unaware K-fold cross-validation setup.

The primary architectural difference between TabPFN and TabICL lies in their attention mechanisms: TabICL employs a column-then-row attention scheme, which enables it to operate faster than TabPFN -- up to 10 times as per \cite{qu-arxiv-tabicl}. Since TabICL outperformed TabPFN in the \ac{ICL} setup, it is reasonable to expect that fine-tuning TabICL could further improve performance, which we leave for future work due to the unavailability of pre-trained TabICL checkpoints.

% It is worthwhile to note that the partial privacy setting can potentially be extended to \ac{TFM} ..
% % Earlier limitation which is no longer valid because  that is now considered partial privacy
% We demonstrate the use of \acp{TFM} on statistical attributes calculated from temporal video series. While effective, this approach may limit the model's ability to capture patterns present in the raw temporal data. Incorporating the full temporal sequence into the \ac{TFM} could potentially reveal additional patterns and further improve classification performance.

\textit{Privacy-utility trade-off.} Comparing the two regimes in \Cref{tab:result-logo}, the partial-privacy Transformer attains higher AUC than strong-privacy TFMPathy ($0.827$ vs $0.669$), while accuracy remains comparable ($0.746$ vs $0.730$). This is the trade-off a data custodian must weigh: temporal features yield higher ranking-based utility but retain information that supports face reconstruction and re-identification \citep{di2018gpgan,kroger2020what}.

\begin{figure}[t!]
    \centering
    \includegraphics[width=\columnwidth]{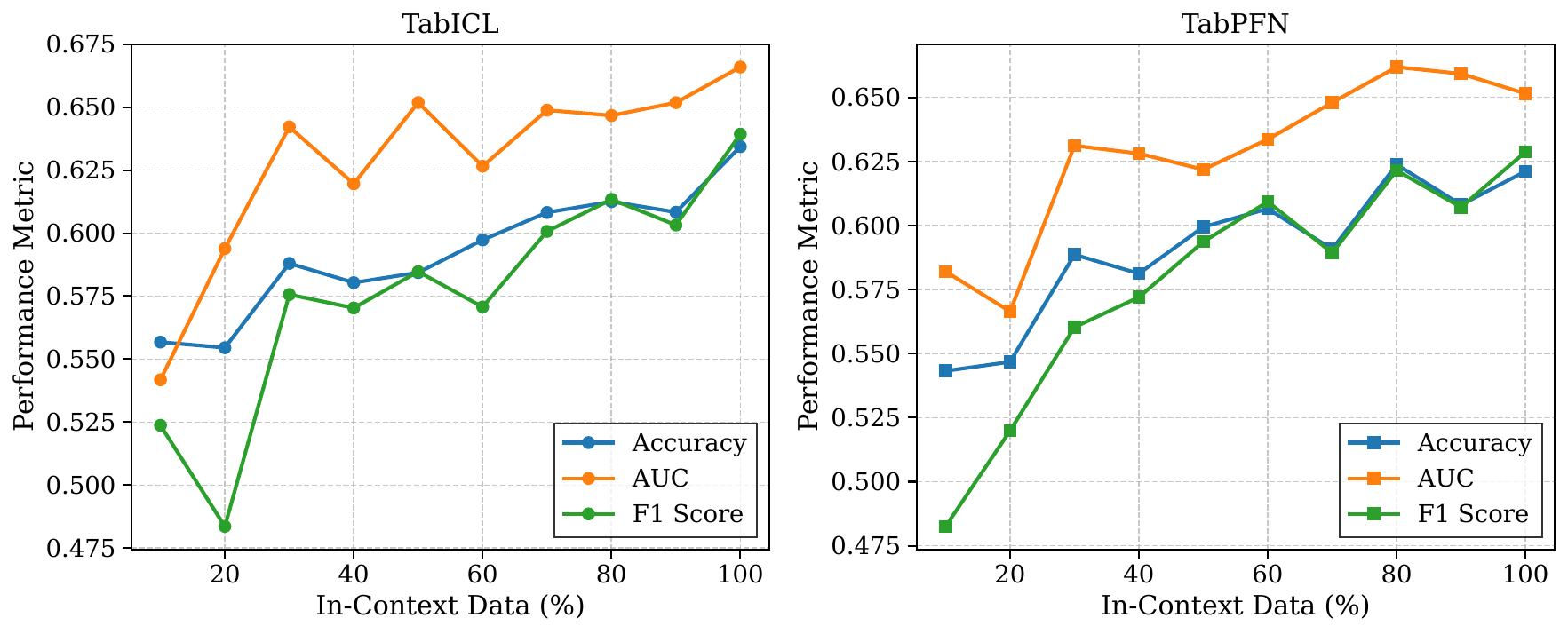}
    \caption{\acf{ICL} performance of \acs{TFMPathy} across different proportions of in-context data (\ie{} training data used in the single forward pass of \acl{ICL}). Results are based on 5-fold, 10-repeat stratified cross-validation without subject group separation.}
    \label{fig:scaling}
\end{figure}

\subsection{Ablation Study}\label{subsec:ablation}
\Cref{tab:tabpfn-ablation} reports a series of experiments on fine-tuning TabPFN (\acs{TFMPathy}\textsubscript{FT}). Key hyperparameters explored include learning rate, batch size and optimiser choice. We experimented with the recent scheduler-free optimiser \citep{defazio2024road} and the standard AdamW optimiser, with and without a learning rate scheduler. Overall, TabPFN showed minimal performance variation across optimiser choices, with AdamW and the one-cycle scheduler yielding the best results. Like optimiser choice, batch size has a modest effect on performance. Increasing it from 8 to 32 improved classification accuracy from 0.701 to 0.708 and \ac{AUC} from 0.689 to 0.696. 

We also tested freezing three pre-trained layers of TabPFN: the X encoder (768 trainable parameters), the Y encoder (576 trainable parameters), and the Transformer encoder (7.1 million trainable parameters). Freezing the X and Y encoders led to improved performance, presumably because they are already well-generalised from TabPFN's large-scale pre-training. Freezing preserves generalisable representations and enables faster fine-tuning, so we kept the X and Y encoders frozen and kept the remaining layers trainable.

We further varied the softmax temperature and the number of input features. Using 500 features led to a slight performance improvement. A SoftMax temperature of 1.0 (\ie{} no temperature scaling) yielded the best classification accuracy (\Cref{tab:tabpfn-ablation}). Based on these findings, the final model uses the AdamW optimiser with a one-cycle scheduler, a batch size of 32, frozen X and Y encoders, a softmax temperature of 1.0, and 500 selected input features.

\Cref{fig:scaling} shows the \ac{ICL} performance of the \acs{TFMPathy} system using both TabICL and TabPFN across varying amounts of context data. In the \ac{ICL} setup, the context data -- \ie{} training data -- is fed into the model during the single forward pass. Both models generally exhibit improved performance as the amount of context data increases, which suggests the potential for further gains with additional data. However, TabPFN's \ac{AUC} peaked when using 80\% of the available context data.

% \missingfigure{Probably: a figure to show how the fine-tuning scales with the number of training data}

\subsection{Limitations \& Future Work}
While our strong-privacy approach reduce privacy risk through summary statistics, it does not provide a formal privacy guarantee.
Our fairness evaluation is limited to individual level via cross-subject generalisation due to unavailability of demographic annotations in the HRI benchmark. Direct demographic subgroup-level fairness evaluation on annotated benchmarks is a meaningful next step. We were similarly unable to fine-tune TabICL due to the absence of pre-trained checkpoints, although its strong in-context performance suggests doing so would likely improve results further.

% The HRI benchmark itself impose additional limitations. 
% The dataset is small (122 samples from 42 subjects) and 
Empathy labels in the HRI dataset are self-reported, which aims to capture the listener's internal experience but may introduce subjective noise. The original authors binarised these labels using the median empathy score, which is a reasonable choice given the sample size but discards information about label magnitude and may obscure cases near the decision boundary. Finally, future work could validate our approach to other external datasets to further strengthen the generalisability of our findings.
% no raw-video baseline possible?

\section{Conclusion}
This paper proposes \acs{TFMPathy}, a privacy-preserving framework that uses summary features from interaction videos in \acfp{TFM} to detect empathy elicited through narrative storytelling. Raw video sequences are likely to breach privacy and raise ethical constraints, so they are often unavailable in public benchmarks. \acs{TFMPathy} learns from visual features and does not require raw video sequences. Experimented on two recent \acp{TFM} (TabICL and TabPFN), we evaluate both \ac{ICL} and fine-tuning approaches to ensure strong privacy. In a public evaluation benchmark on human-robot interaction, \acs{TFMPathy} in both \ac{ICL} and fine-tuning settings outperforms several baselines while maintaining better privacy than sequence learning methods. In addition to the established protocol, we evaluate cross-subject generalisation in detecting empathy elicited through storytelling. While most models, including \acp{TFM} in the initial \ac{ICL} setup, struggle in this evaluation protocol, \acp{TFM} in the fine-tuning setup achieve substantially better generalisation across unseen subjects. Given the persistent privacy and ethical constraints in raw videos, our proposed \acs{TFMPathy} system offers practical value and future potential for AI systems that rely on human-centred video datasets.

\section*{Acknowledgements}
This work was supported by resources provided by the Pawsey Supercomputing Research Centre with funding from the Australian Government and the Government of Western Australia. We thank Dr Md Redowan Mahmud of Curtin University for reviewing our manuscript and for his insightful comments, including those on privacy.

% {\appendix[Proof of the Zonklar Equations]
% Use $\backslash${\tt{appendix}} if you have a single appendix:
% Do not use $\backslash${\tt{section}} anymore after $\backslash${\tt{appendix}}, only $\backslash${\tt{section*}}.
% If you have multiple appendixes use $\backslash${\tt{appendices}} then use $\backslash${\tt{section}} to start each appendix.
% You must declare a $\backslash${\tt{section}} before using any $\backslash${\tt{subsection}} or using $\backslash${\tt{label}} ($\backslash${\tt{appendices}} by itself
%  starts a section numbered zero.)}


\printbibliography

% \newpage

 

% \end{IEEEbiography}



% \vfill

\begin{acronym}
    \acro{AUC}{Area Under the receiver operating characteristics Curve}
    \acro{GPT}{Generative Pre-trained Transformer}
    \acro{HRI}{Human-Robot Interaction}
    \acro{ICL}{In-Context Learning}
    \acro{LLM}{Large Language Model}
    \acro{LOSO}{Leave-One-Subject-Out}
    \acro{LVM}{Large Vision Model}
    \acro{LSTM}{Long Short-Term Memory}
    \acro{ML}{Machine Learning}
    \acro{PFN}{Prior-data Fitted Network}
    \acro{RF}{Random Forest}
    \acro{SVM}{Support Vector Machine}
    \acro{SOTA}{State-of-the-Art}
    \acro{TCN}{Temporal Convolutional Network}
    \acro{TFM}{Tabular Foundation Model}
    \acro{TFMPathy}{\ac{TFM} for Empathy}
\end{acronym}

\end{document}